\documentclass{article}

\usepackage{PRIMEarxiv}

\usepackage[utf8]{inputenc} 
\usepackage[T1]{fontenc}    
\usepackage{hyperref}       
\usepackage{url}            
\usepackage{booktabs}       
\usepackage{amsfonts}       
\usepackage{nicefrac}       
\usepackage{microtype}      
\usepackage{lipsum}
\usepackage{fancyhdr}       
\usepackage{graphicx}       
\graphicspath{{media/}}     
\usepackage{amssymb}
\usepackage{amsmath}
\usepackage{subcaption}
\usepackage{algorithm}
\usepackage{algpseudocode}
\title{Parsing the Language of Expression: Enhancing Symbolic Regression with Domain-Aware Symbolic Priors
}

\author{
Sikai Huang \\
  Department of Mathematics \\
  Purdue University \\
  West Lafayette, IN, USA\\
  \texttt{huan1580@purdue.edu} \\
   \And
  Yixin Berry Wen \\
  Department of Geography \\
  University of Florida \\
  Gainesville, FL, USA\\
  \texttt{yixin.wen@ufl.edu } \\
   \And
\And
Tara Adusumilli, Kusum Choudhary \\
 Department of Computer Science \\
University of Maryland College Park \\
College Park\\
  \texttt{\{tadusumi, kusumcho\}@umd.edu} \\
   \And
Haizhao Yang \\
 Department of Mathematics, Department of Computer Science \\
University of Maryland College Park \\
College Park\\
  \texttt{hzyang@umd.edu} \\
}

\begin{document}
\maketitle

\begin{abstract}
Symbolic regression is essential for deriving interpretable expressions that elucidate complex phenomena by exposing the underlying mathematical and physical relationships in data. In this paper, we present an advanced symbolic regression method that integrates symbol priors from diverse scientific domains - including physics, biology, chemistry, and engineering - into the regression process. By systematically analyzing domain-specific expressions, we derive probability distributions of symbols to guide expression generation. We propose novel tree-structured recurrent neural networks (RNNs) that leverage these symbol priors, enabling domain knowledge to steer the learning process. Additionally, we introduce a hierarchical tree structure for representing expressions, where unary and binary operators are organized to facilitate more efficient learning. To further accelerate training, we compile characteristic expression blocks from each domain and include them in the operator dictionary, providing relevant building blocks. Experimental results demonstrate that leveraging symbol priors significantly enhances the performance of symbolic regression, resulting in faster convergence and higher accuracy.
\end{abstract}


\section{Introduction}
\subsection{Problem Statement}
Symbolic regression is a powerful technique that searches the space of mathematical expressions to identify equations that best fit a dataset. Unlike traditional regression models that rely on predefined structures, symbolic regression discovers interpretable relationships between variables, offering deeper insights into data dynamics. This capability is particularly valuable in fields with complex, poorly understood relationships in various fields, e.g., physical sciences \cite{angelis2023artificial, miles2021machine, neumann2020new}, materials science \cite{wang2019symbolic, wang2022symbolic}, chemistry \cite{neumann2020new, xie2022machine, hu2023first}, climate science, ecology, \cite{chen2019revealing, martin2018reverse, cardoso2020automated}, and finance \cite{duffy2002using, jin2019bayesian}. These diverse applications underscore the versatility of symbolic regression as a powerful tool for scientific discovery and analysis.

Symbolic regression methods are generally categorized into two main approaches. The first approach involves an optimization process to identify suitable expressions. This can be a two-step process: first, generating a ``skeleton" of the equation using a parametric function built from a predefined set of operators according to physical knowledge, such as basic arithmetic operations and elementary functions, to define its overall structure. The second step then solves a regression problem to estimate parameters within this skeleton. Alternatively, both the function skeleton and parameters can be learned simultaneously through mixed optimization. This problem is typically addressed using genetic algorithms \cite{blkadek2019solving, schmidt2009distilling, mirjalili2019genetic, langdon2013foundations, mundhenk2021symbolic} or, more recently, reinforcement learning (RL) \cite{petersen2019deep, mundhenk2021symbolic, liang2022finite, sun2022symbolic}.

Inspired by recent advances in language models, a second approach to symbolic regression has emerged as an end-to-end solver, often referred to as Neural Symbolic Regression (NSR). This approach frames symbolic regression as a natural language processing (NLP) task, leveraging large-scale pre-trained models to map data directly to expressions in an end-to-end manner, akin to machine translation \cite{bendinelli2023controllable, kamienny2022end, vastl2024symformer, shojaee2024transformer, li2022transformer, merler2024context}. These neural models are trained end-to-end, taking sampled data points as input and generating symbolic representations of mathematical expressions that best fit the data.

\subsection{Symbolic Prior for Symbolic Regression} 
Symbolic regression is extensively employed to derive interpretable expressions that characterize dynamical systems in a wide range of scientific domains, including physics, biology, and chemistry. However, the frequency and combination of symbols and operators differ markedly across these fields, reflecting their underlying principles and commonly adopted mathematical formulations. For example, trigonometric functions (e.g., sine and cosine) often appear in physics to capture oscillatory behaviors, whereas exponential and logistic functions are more prevalent in biology to model population growth and decay. These observations naturally lead to two questions:
\\
\par 
\textit{How can we systematically extract these symbol priors? Moreover, how can we efficiently incorporate such prior knowledge to enhance existing symbolic regression methods?}
\\

\begin{figure}[ht]
\begin{center}
\includegraphics[width=6.5in]{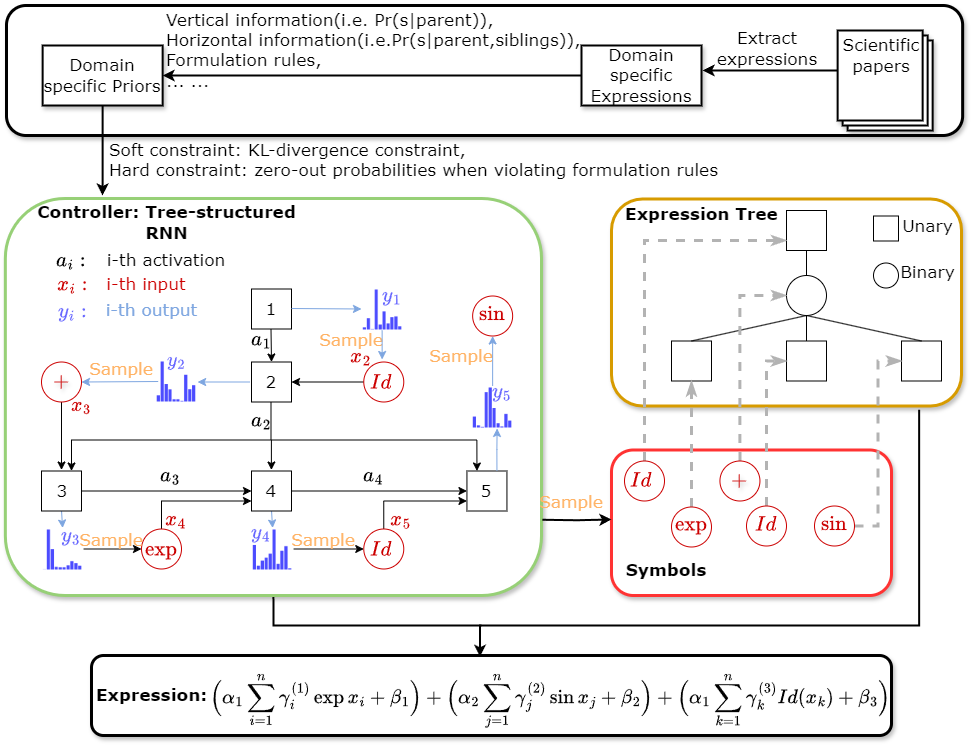}
\end{center}
\caption{A tree-structured RNN-based reinforcement learning framework for generating symbolic expressions. Domain-specific priors (top) are incorporated as soft constraints (KL divergence) and hard constraints (rule-based masking), guiding the controller to propose expressive yet valid “skeletons.” Sampled expressions are then refined by optimizing their parameters, yielding interpretable mathematical models aligned with target data.}
\label{fig:rl_framework}
\end{figure}

\begin{figure}[ht]
\begin{center}
\includegraphics[width=6in]{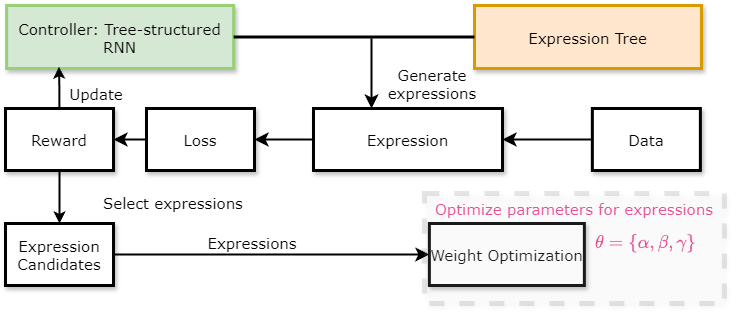}
\end{center}
\caption{Parameter optimization and candidate selection loop. The tree-structured RNN controller generates candidate expressions, which are evaluated against the data to produce a reward signal. High-scoring candidates are selected for further  \emph{parameter tuning}, where the weights $\theta=\{\alpha, \beta, \gamma\}$ are refined to better fit the target data, ultimately yielding more accurate symbolic expressions.}
\label{fig:opt_process}
\end{figure}

The main contributions of this paper for the above questions are summarized below.
\begin{itemize}
    \item [$\square$] \textbf{Novel Tree Representation of Expressions: } This work proposes a general (multi-branch) tree representation for mathematical expressions, effectively capturing hierarchical structures, particularly in consecutive additions. By treating linked unary operators as equivalent nodes, the method preserves local structures that binary trees and linear sequences often fail to maintain due to increased depth and imbalance. The output of a leaf node is expressed as a linear combination of variables applied element-wise to the same unary operator, reducing tree depth and yielding a more compact expression.


    \item [$\square$]
    \textbf{Collection and Categorization of Domain-Specific Expressions.} This study systematically gathers mathematical expressions from relevant arXiv papers and organizes them by scientific domain. Leveraging our general tree-structure representation, we analyze domain-specific symbol relationships and operator combinations to extract priors and refine the operator set, thereby improving training efficiency. We classify these priors into two categories: \emph{horizontal priors}, which capture relationships among sibling unary operators under the same parent, and \emph{vertical priors}, which characterize hierarchical dependencies between a node and its ancestors. Conditional categorical distributions encode these intrinsic horizontal and vertical features, thus providing a domain-aware understanding of expression patterns.



    \item [$\square$] \textbf{Tree-Structured RNN Policy Optimized with KL Regularization: } 
As illustrated in Figure~\ref{fig:rl_framework}, we employ a \emph{tree-structured} recurrent neural network (RNN) to represent the policy within a reinforcement learning (RL) framework for generating mathematical expressions. This hierarchical design leverages the nested structure of expressions, thereby reducing the number of RNN modules required for capturing complex operator interactions. To incorporate \emph{domain-specific} symbol priors, we introduce a Kullback-Leibler (KL) divergence regularization term into the reward function, effectively minimizing the discrepancy between the policy’s learned distribution and the prior distribution. The policy is trained via policy gradient methods, exploring the space of symbolic expressions and maintaining a pool of high-scoring “skeletons.” As depicted in Figure~\ref{fig:opt_process}, these candidate expressions are iteratively refined to converge toward the target equation.

\end{itemize}

By incorporating domain-specific symbol priors into the hierarchical RNN’s training procedure, the model leverages relevant prior knowledge to enhance both the efficiency and accuracy of symbolic regression. Empirical evaluations indicate that this strategy not only accelerates convergence but also yields more accurate and interpretable models across diverse scientific domains.

\subsection{Related work}
Numerous strategies have been proposed to incorporate prior knowledge into symbolic regression. One prevalent approach involves balancing interpretability and data fidelity through complexity-based penalties, such as the Bayesian Information Criterion (BIC) \cite{bastiani2024complexity}. Another line of research assigns priors to tree structures and operators via uniform distributions or user-defined preferences, yet lacks a systematic way to derive domain-specific priors \cite{jin2019bayesian}. In contrast, the method introduced here employs \emph{domain-specific} symbol priors to improve both accuracy and interpretability. By adding a Kullback--Leibler (KL) divergence term to the reward function, it aligns the learned categorical distribution with domain knowledge, thereby achieving more robust control and faster convergence relative to existing techniques.

To further constrain the search space, several works have imposed \emph{structural} constraints on expressions. For instance, monotonicity, convexity, and symmetry constraints have been incorporated to guide the search, thereby improving efficiency \cite{gupta2016monotonic,bezerra2019archiver,kronberger2022shape}. Although effective, these constraints can be difficult to infer from data alone and may not always be applicable. Similarly, research on embedding fundamental physical laws has used conservation principles to enforce physical validity \cite{ashok2021logic,kubalik2021multi}, but such methods typically require detailed \emph{a priori} knowledge about the system.

Beyond structural constraints, some approaches leverage optimization algorithms, notably genetic algorithms, to validate candidate solutions under predefined conditions such as symmetry, monotonicity, convexity, and boundary constraints \cite{blkadek2022counterexample}. However, these strategies often rely on data quality for verifying constraints, which can be challenging in practice. Alternatively, context-sensitive filtering methods selectively prune unlikely token sequences by examining the structure of the expression tree, thereby reducing invalid symbol combinations and enhancing efficiency \cite{petersen2019deep,petersen2021incorporating}.

A further avenue for integrating prior knowledge involves \emph{domain-specific} constraints on symbolic representations. For example, dimensional consistency has been enforced by masking the categorical distribution in RNNs, ensuring physically meaningful expressions \cite{tenachi2023deep}. Building on this idea, our approach introduces ``hard constraints" that rule out symbol combinations not observed in specific domains, supplemented by probabilistic biases on token combinations to further refine the search. This combination restricts the solution space to expressions that are both more plausible and interpretable within the given domain.

\section{Symbol Priors}
In this section, we present a strategy for incorporating symbol priors into symbolic regression. We begin by introducing a \emph{tree-structured} representation of mathematical expressions that systematically gathers and utilizes symbol priors, enabling a more compact and structured encoding of expressions for cross-domain analysis. We then describe a method for extracting symbol priors from domain-specific mathematical expressions in arXiv papers. By capturing the characteristic symbol distributions and operator preferences of each discipline, this approach leverages structured domain knowledge to enhance both the accuracy and interpretability of symbolic regression.


\subsection{Representation Method}

Our proposed representation addresses the shortcomings of conventional binary expression trees by permitting a single binary operator to connect multiple sequences of unary operators. This multi-branch design yields a more flexible and expressive hierarchical structure. Empirical analysis of real-world expressions, particularly those describing dynamical systems, shows that most physically meaningful forms can be effectively captured within a two-level tree structure—some even reduce to a single layer. By aligning closely with these practical mathematical constructs, the representation improves interpretability and boosts the efficiency of symbolic regression.
\begin{figure}[h!]
\begin{center}
\includegraphics[width=6in]{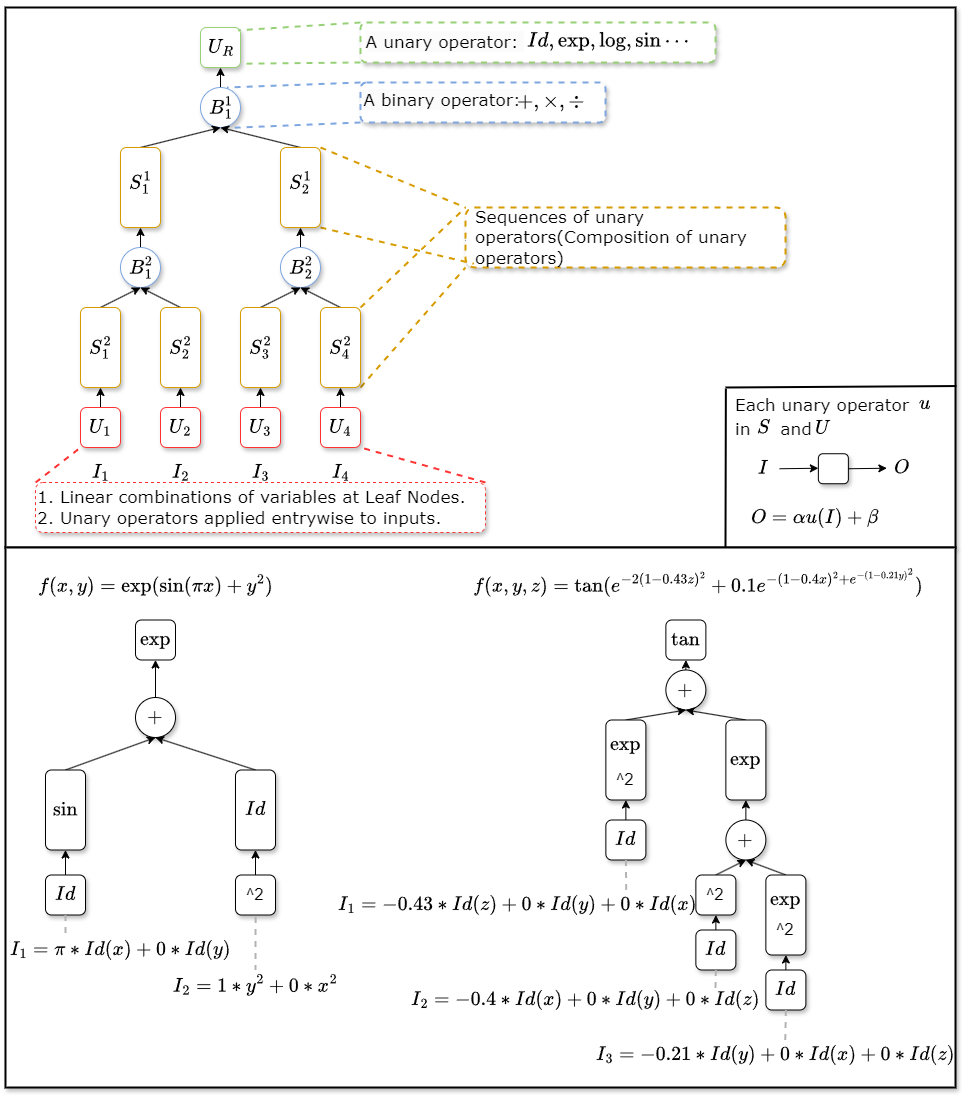}
\end{center}
\caption{The top panel illustrates the fundamental structure of our representation method, while the bottom panel presents two example expressions represented using this structure.
}
\label{fig: 3}
\end{figure}

For instance, in the case of consecutive additions, one addition operator can connect multiple child nodes that are treated equivalently, thus eliminating strict hierarchical dependencies often found in binary trees. By contrast, traditional binary representations impose tiered parent–child relationships, leading to deeper and more inflexible structures.

To formalize our representation method, we define the following sets:

\begin{itemize}
    \item [$\bullet$]\textbf{Unary Operator Set}: Let $\mathcal{U} = \{\sin, \exp, \log, Id, (\cdot)^2, \dots\}$, which encompasses a range of elementary functions (e.g., polynomials, trigonometric functions). Here, $Id$ denotes the identity function.

    \item [$\bullet$]\textbf{Binary Operator Set}: Let \(\mathcal{B} = \{+, \times, \div\}\) represent the set of binary operators within the tree structure.

    \item [$\bullet$]\textbf{Variable Set}: 
    Let \(V = \{ f, x_1, \dots, x_n, f_{x_i}, f_{x_i x_j} \,\mid\, 1 \leq i, j \leq n\}\), 
where \(f\) is the primary function, \(x_1, \dots, x_n\) are variables, \(f_{x_i} = \frac{\partial f}{\partial x_i}\) denotes the \emph{first-order partial derivative} of \(f\), and \(f_{x_i x_j} = \frac{\partial^2 f}{\partial x_i \partial x_j}\) denotes the \emph{second-order partial derivative}. Higher-order derivatives are typically uncommon in most physical systems.
\end{itemize}
With these sets established, our representation method—illustrated in Figure~\ref{fig: 3}—integrates unary operators (and their compositions) with binary operators in a hierarchical framework. Specifically:
\begin{itemize}
    \item [$\bullet$] \textbf{Root Node ($U_{R}$)}: 
    The root node is a unary operator chosen from $\mathcal{U}$. It applies its operation to the output of its subtrees, which are connected by a binary operator from $\mathcal{B}$.

   \item [$\bullet$] \textbf{Root Node Binary Operator Connection ($B$)}: A binary operator from \(\mathcal{B} = \{+,\,\times,\,\div\}\) links multiple sequences of unary operators. For example, \(B^1\) connects the first-level sequences \(S_1^1\) and \(S_2^1\) to the root node \(U_R\), merging sub-expressions without enforcing the strict parent--child hierarchy typical of standard binary trees.


    \item [$\bullet$] \textbf{Sequences of Unary Operators ($S_{i}^{j}$)}: Each \(S_i^j\) is a composition of unary operators drawn from \(\mathcal{U}\) associated with \(B_{\lfloor (i+1)/2\rfloor}^j\). In particular, \(S_i^1\) denotes the \emph{first-level} sequences associated with \(B_1^1\), while \(S_i^2\) denotes the \emph{second-level} sequences linked by \(B_i^2\). By allowing multiple unary operators to be chained, this structure captures a broader range of real-world expressions.

    \item [$\bullet$] \textbf{Leaf Nodes ($U_{i}$)}: The leaf nodes, denoted by \(I_i\), take inputs from the variable set \(V\), which includes the function \(f\), its partial derivatives, and the variables \(x_1, \dots, x_n\). Each leaf applies a unary operator \emph{element-wise} to these inputs and outputs a linear combination of the resulting transformed variables:
\[
O \;=\; \gamma_1\,\mu(v_1)\;+\;\gamma_2\,\mu(v_2)\;+\;\dots\;+\;\gamma_n\,\mu(v_n), 
\quad v_i \,\in\, V,
\]
where \(\mu(\cdot)\) is the chosen unary operator and \(\gamma_i\) are learned coefficients.
\item[\(\bullet\)] \textbf{Linear Transformation in Non-Leaf Unary Operators:}
Each non-leaf unary operator \(\mu\) in \(U_R\) or in any \(S_i^j\) undergoes a linear transformation:
\[
O \;=\;\alpha\,\mu(I)\;+\;\beta,
\]
where \(\alpha\) is a scaling factor and \(\beta\) is a bias term. This additional affine component expands the representational flexibility for modeling complex functional relationships.

\end{itemize}
\textbf{Remark. }
The coefficients associated with variables in each leaf node’s linear combination can be optimized as part of the model-fitting process. This strategy implicitly performs \emph{feature selection}, allowing the discovery of the most pertinent variables and revealing core relationships governing the system.
\noindent

\par
Within our proposed framework, we introduce three fundamental concepts pertaining to each expression: \emph{subsequences}, \emph{width}, and \emph{depth}. Formal definitions and illustrative examples are given as follows.

Let \(\mathcal{T}\) be an \(h\)-level expression tree with a unique root unary operator \(U_R\) 
and leaf nodes \(\{I_k\}_{k=1}^n\), where each leaf node \(I_k\) is associated with a unary 
operator \(U_k\).

\noindent
\textbf{Definition 1 (Subsequence of an Expression). } 
A \emph{subsequence} for a particular leaf node \(I_k\) is defined as the 
ordered list of unary and binary operators encountered along the unique path from the root 
node \(U_R\) down to \(U_k\). If at each level \(\ell \in \{1, \dots, h\}\), the path may include a binary operator \(B^\ell\) 
and an associated sequence of unary operators \(S^\ell\), then formally:
\[
\mathrm{Subsequence}(I_k) 
\,=\, 
\bigl\{\,U_R,\;B^1,\;S^1,\;B^2,\;S^2,\;\dots,\;B^h,\;S^h,\;U_k\bigr\},
\quad
k \in \{1,\dots,n\},
\]
where \(U_R\) is the root unary operator, \(B^1,\ldots,B^h\) are the binary operators 
encountered along the path, \(S^1,\ldots,S^h\) are the corresponding unary-operator 
sequences, and \(U_k\) is the final unary operator at leaf node \(I_k\). In cases where 
fewer than \(h\) levels are traversed, the omitted operators are simply excluded from 
the subsequence.



\noindent
\textbf{Definition 2 (Width of an Expression). }
The \textit{width of an expression} is defined as the total number of first-level sequences. Formally, given a tree $\mathcal{T}$ with outermost unary operator sequences $\{S_1^1, S_2^1, \dots, S_m^1\}$, the width of $\mathcal{T}$ is 
    \[\text{Width}(\mathcal{T}) = m,\] 
where $m$ denotes the number of unary-operator sequences directly connected to the root node.

\noindent
\textbf{Definition 3 (Depth of an Expression). }
The \textit{depth} of an expression is defined as the length of the longest subsequence in the expression tree $\mathcal{T}$. Equivalently, it is the greatest number of unary operators on any path from a leaf node $I_k$ to the root node $U_R$. Formally,

   \[
   \text{Depth}(\mathcal{T}) = \max_{1 \leq k \leq n} \left( \text{Length of Subsequence}(I_k) \right),
   \]
   where the length of the subsequence is the total number of unary operators encountered between $U_{R}$ and $I_k$ and $n$ is the total number leaf nodes.
\begin{figure}[ht]
\begin{center}
\includegraphics[width=6.6in]{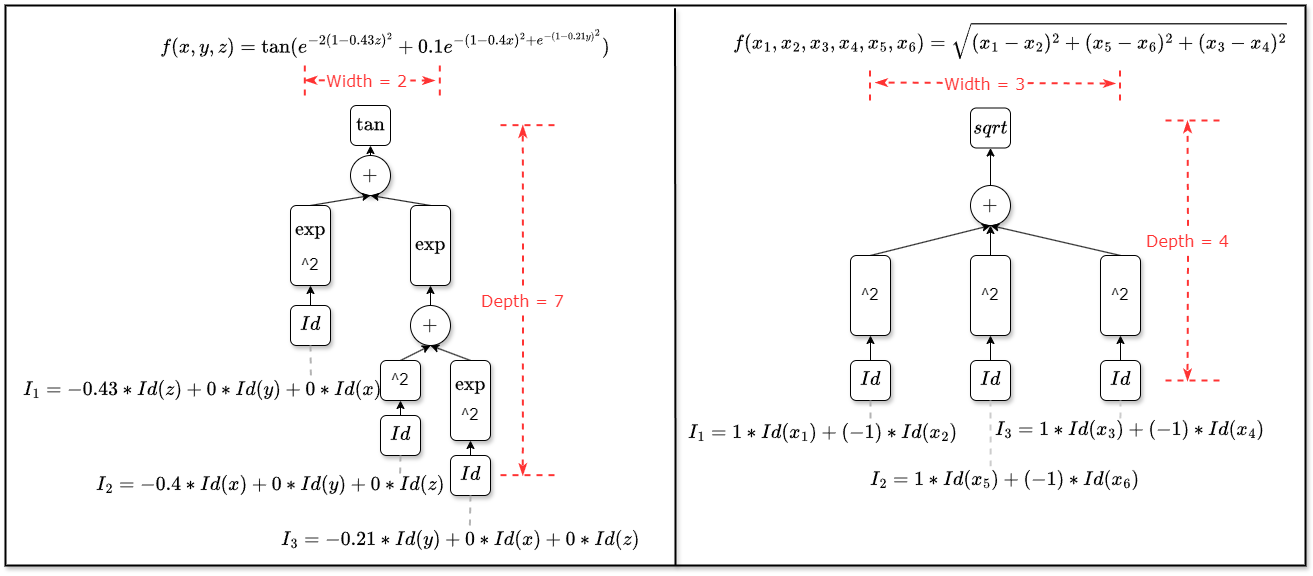}
\end{center}
\caption{Examples of two expression trees, illustrating their subsequences, width, and depth.
}
\label{fig: 15}
\end{figure}
Figure 4 presents two illustrative examples explained below:
\begin{itemize}
    \item [$\bullet$] \textbf{Left example}: The subsequences are $\{ \tan, +, \exp, (\cdot)^2, \text{Id} \}$, $\{ \tan, +, \exp, +, (\cdot)^2, \text{Id} \}$, and $\{ \tan, +, \exp, +, \exp, (\cdot)^2, \text{Id} \}$. Since there are two first-level sequences connected directly to the root node, the \emph{width} of the tree is 2. The \emph{depth} is 7, reflecting the length of the longest path from the root to a leaf node.
    \item [$\bullet$] \textbf{Right example}: This tree contains three subsequences of the form \(\{ \sqrt, +, (\cdot)^2, \text{Id} \}\). Because three first-level sequences directly connect to the root node, the \emph{width} of the tree is 3. Its \emph{depth} is 4, corresponding to the length of the longest path from a leaf node to the root.
\end{itemize}
To maintain consistency in our representation, the identity operator \textbf{Id} is allowed in the root node, leaf nodes, and any sequence $S$. However, within a given sequence $S$, \textbf{Id} is permitted only if it is the \emph{sole} unary operator. Superfluous occurrences of \textbf{Id} increase the overall expression length and hinder subsequent symbol-prior extraction. Moreover, restricting \textbf{Id} at the leaf nodes helps preserve a concise and interpretable representation.

\subsection{Hierarchical Symbol Priors Extraction}
We systematically gathered mathematical expressions from arXiv, targeting specific scientific domains. For each domain, 10{,}000 highly relevant papers were selected, and their embedded expressions were extracted to examine structural patterns critical to our approach. Each expression was then converted into a general tree structure, capturing elements such as subsequences, root nodes, root-level binary operators, leaf nodes, as well as the tree’s width and depth. This structured format enables a detailed analysis of symbol relationships and domain-specific usage. Equipped with this extensive repository of expressions and subsequences, we proceed to extract the following core information.

\textbf{Vertical Tree Node Analysis:}
The goal is to capture vertical compositional operator relationships from root to leaf nodes. By examining each path, we estimate conditional categorical distributions of symbols at multiple levels, thus deriving \emph{vertical} symbol priors that encapsulate domain-specific operator patterns.

Notably, numerous symbol combinations occur with zero probability. In many cases, these combinations violate the \textit{General Formulation Rules} \cite{petersen2019deep}, which impose restrictions such as limiting trigonometric nesting to two levels (e.g., disallowing $\cos(x+\sin(y+\tan))$), restricting self-nesting of exponential and logarithmic functions (e.g., $\exp(\exp(\cdot))$, $\log(\log(\cdot))$), and forbidding direct succession of inverse unary operations (e.g., $\exp(\log(\cdot))$, $\log(\exp(\cdot))$).


\textbf{Horizontal Tree Node Analysis: }
From a horizontal perspective, we investigate sibling nodes connected by a common binary operator at each tree level. Specifically, for each level $h$, we aggregate all child nodes linked via the same operator $B$ and estimate categorical distributions of the observed symbol combinations. This procedure reveals how operators co-occur laterally at the same level and enables more precise modeling of domain-specific patterns.

\textbf{Domain-Specific Component Analysis: }
Note that certain substructures, composed of both unary and binary operators, appear frequently within specific domains. In engineering—particularly in signal processing—combinations such as $\cos(\cdot) + \sin(\cdot)$ often model waveforms, whereas in chemistry, exponential constructs like $\exp(\cdot / \cdot)$ frequently arise in reaction-rate formulations (e.g., the Arrhenius equation for temperature-dependent reaction rates). Incorporating these domain-specific components in the expression search can significantly improve the computational efficiency of symbolic regression.



\textbf{Other Priors:} 
Beyond specific operator combinations, we also extract prior information from each expression tree, including distributions of root nodes, leaf nodes, and structural characteristics such as depth and width. The root node determines the overarching form of the expression, while leaf nodes serve as anchors for variables or constants. Examining these distributions uncovers domain-specific preferences for certain functions and operations. Furthermore, structural priors involving depth and width help control the complexity of candidate expressions, preventing solutions that are either overly simplistic or excessively convoluted.\\
\textbf{Remark:}  
The total number of variables within an expression affects the raw frequency counts of operator combinations, potentially causing statistical bias. For instance, consider the expression \(\sum_{i=1}^{n}\cos(x_i)\). As the number of variables \(n\) increases, the raw count of the unary operator \(\cos(\cdot)\) increases proportionally, artificially inflating its frequency relative to other operators. This bias necessitates an appropriate normalization method to accurately reflect the true operator distributions across diverse expressions.

\noindent
\textbf{Definition 4 (Normalization by Variable Count).} Let \(\mathcal{E}\) denote the corpus of collected expressions. To address the bias introduced by varying numbers of variables across different expressions, we define a normalized frequency measure for operator combinations across the corpus. Specifically, for an operator combination \( s \), the normalized count within a single expression \( E \in \mathcal{E} \) is computed as:
\[
\text{Normalized Count}_E(s) = \frac{\text{Raw count of operator combination } s \text{ in expression } E}{\text{Total number of variables in expression } E}.
\]

The overall normalized frequency for the operator combination \( s \) is then defined by averaging these normalized counts over all expressions in the corpus \( \mathcal{E} \):
\[
\text{Normalized Count}(s) = \frac{1}{|\mathcal{E}|}\sum_{E \in \mathcal{E}} \text{Normalized Count}_E(s).
\]

This approach yields a balanced and representative probability distribution of operator combinations across diverse expressions.

\noindent
\textbf{Definition 5 (Prior Probability for Node Symbols).} Let \(s_i\) be a node in an expression tree with parent node \(s_p(h)\) at level \(h\). Denote by \(S_i\) the set of sibling symbols of \(s_i\). We define the \emph{prior probability} of \(s_i\) given its siblings \(S_i\), parent \(s_p\), and level \(h\) as follows:

\begin{itemize}
\item \textbf{If the number of siblings \(|S_i| \leq 3\)}, we explicitly consider all siblings:

\[
S_i^* = (s_{i-1},\, s_{i-2},\, s_{i-3}), \quad \text{with fewer symbols included if fewer siblings exist.}
\]

\item \textbf{If the number of siblings \(|S_i| > 3\)}, we consider only the three most frequently occurring siblings \(s_{(1)}, s_{(2)}, s_{(3)}\):

\[
S_i^* = (s_{(1)},\, s_{(2)},\, s_{(3)}).
\]
\end{itemize}

Using normalized counts as defined earlier, the prior probability is rigorously estimated by aggregating normalized frequencies across a corpus \(\mathcal{E}\) of multiple expressions:

\[
P^{*}(s_i \mid S_i, s_p(h)) = \frac{\sum_{E \in \mathcal{E}} \text{Normalized Count}_E(s_i, S_i^*, s_p(h))}{\sum_{E \in \mathcal{E}} \text{Normalized Count}_E(S_i^*, s_p(h))},
\]

where \(\text{Normalized Count}_E(\cdot)\) denotes the normalized count computed within an individual expression \(E\).\\

\noindent
\textbf{Remark (Rationale for Restricting Sibling Context).}  
Restricting the sibling context to the three most frequent siblings provides a practical balance between capturing essential statistical information and mitigating combinatorial complexity. Specifically, considering all sibling nodes would result in exponential growth of the combinatorial search space, severely exacerbating data sparsity and estimation instability. Empirical evidence from symbolic regression and related probabilistic modeling tasks suggests that the majority of the predictive information (measured in terms of conditional mutual information) is typically concentrated in a limited subset of frequently co-occurring sibling nodes. Thus, incorporating more than three sibling symbols typically yields diminishing marginal returns in information content, while greatly increasing complexity and reducing statistical robustness.

\subsection{Case Study}
This section systematically compares the horizontal and vertical priors previously outlined, along with distributions of root and leaf nodes and fundamental structural attributes. By examining both individual and combined impacts of these priors, we illustrate how incorporating domain-specific knowledge significantly enhances symbolic regression performance and informs optimal learning framework configurations.

Across all four domains—physics, chemistry, biology, and engineering—expressions with substantial depth are relatively uncommon. The introduction of identity operators, whenever no unary operator exists between consecutive binary operators, increases the effective depth, thus often resulting in actual depths exceeding initial assessments.

Expressions in physics and engineering typically exhibit greater depth due to the nested functions and layered operations necessary for modeling complex phenomena. Specifically, physics expressions frequently involve nested trigonometric or exponential functions, differential equations, and integrals, while engineering models commonly integrate multiple layers of system dynamics and control mechanisms.

The width of expressions generally clusters around moderate values in all domains. While expressions structured as \(\sum_{i=1}^{n}\) superficially suggest potentially unbounded width due to the variable \(n\), the normalized count definition mitigates this issue by proportionally scaling symbol combinations relative to the total variable count. Thus, normalized widths remain statistically manageable despite variations in \(n\). Broad top-level structures occur frequently across domains, reflecting parallel interactions inherent within these systems: chemical equations summing reactants and products, biological models aggregating multiple genetic or environmental factors, engineering calculations combining parallel impedances, and physical expressions summing over multiple states or particles. Nonetheless, the normalized count method ensures consistent and meaningful statistical comparisons of expression widths.

\begin{figure}[H]
\begin{center}
\includegraphics[width=5.7in]{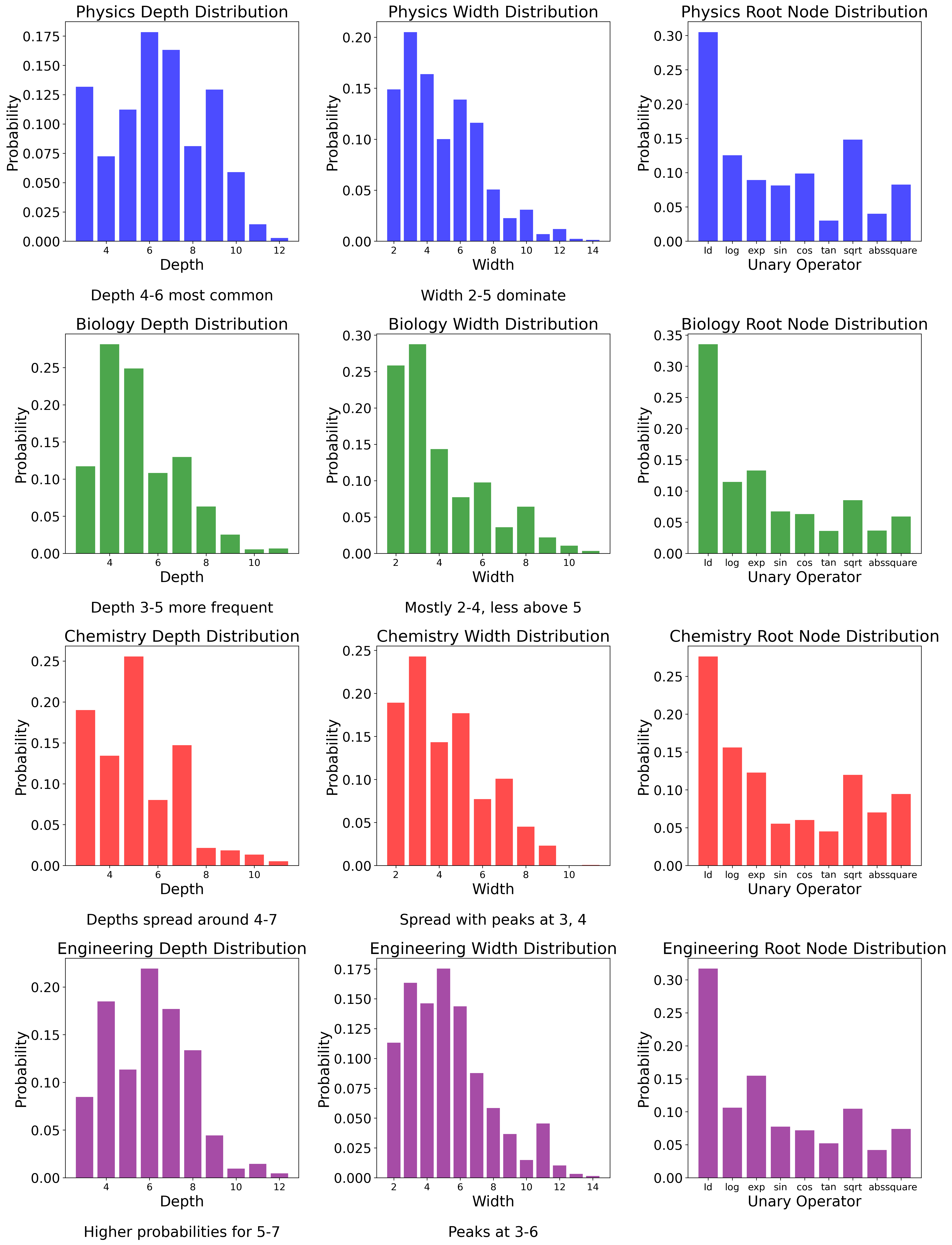}
\end{center}
\caption{Statistical distributions of expression depth, width, and root nodes across Physics, Biology, Chemistry, and Engineering.}
\label{fig: 24}
\end{figure}

Figure~\ref{fig: 24} illustrates the distribution of the binary operator \(B_{1}^1\) conditioned on various root nodes across physics, biology, chemistry, and engineering domains. This vertical analysis reveals how selecting specific root nodes—such as \(\mathrm{Id}\), \(\log\), \(\exp\), \(\sin\), \(\cos\), \(\tan\), \(\sqrt{\cdot}\), or \((\cdot)^2\)—influences subsequent binary operator distributions. By examining these hierarchical dependencies, we gain deeper insight into domain-specific conventions governing the construction of mathematical expressions.

Moreover, the analysis identifies subsequence patterns rarely encountered within these scientific domains, highlighting the practical importance of adhering to established formulation rules. For instance, combinations like \(\sqrt{\log(\tan(\cdot))}\) or \(\sqrt{\tan(\log(\cdot))}\) are infrequently used, as they typically lack clear interpretability or physical relevance, and thus are generally avoided in standard scientific modeling.

\begin{figure}[H]
\begin{center}
\includegraphics[width=6.5in]{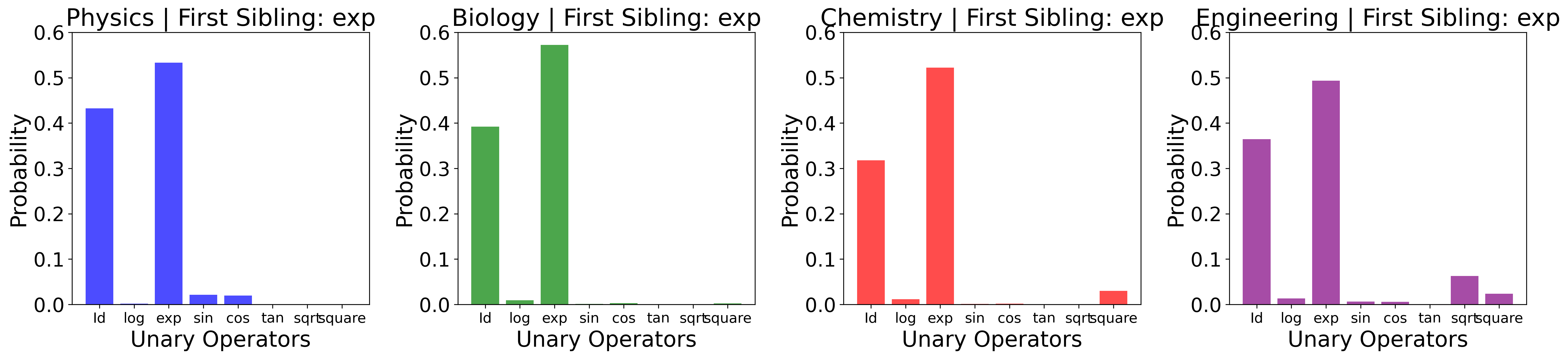}
\end{center}
\caption{Probability Distributions of Second Sibling Unary Operator Given Parent Binary Operator ``+" and First Sibling ``exp" in Various Fields'}
\label{fig: 26}
\end{figure}

Our horizontal analysis focuses on sibling nodes sharing a common binary operator \( B \), examining the conditional distributions of sibling nodes given their parent. Formally, consider a binary parent node \( B \) with sibling child nodes \( s_1 \) and \( s_2 \). Figure~\ref{fig: 26} illustrates empirical conditional distributions of sibling operators across various scientific domains.

Common operand pairs, such as \(\exp + \mathrm{Id}\) and \( \exp + \exp \), frequently arise in all examined fields, representing fundamental models of exponential growth, decay, and the summation of exponentials common in differential equations. In contrast, combinations like \( \exp + \tan \) or \( \exp + (\cdot)^2 \) rarely appear due to their limited physical interpretability and potential numerical instability. Furthermore, domain-specific preferences are evident: physics often employs combinations of exponential and trigonometric functions (e.g., \( \exp + \sin \) or \( \exp + \cos \)) to model oscillatory phenomena; biology typically utilizes simpler forms such as \( \exp + \text{Id} \) or \( \exp + \log \) reflecting fundamental growth dynamics; chemistry frequently combines exponentials with logarithmic functions (e.g., \( \exp + \log \)) due to their role in reaction kinetics; and engineering integrates varied combinations, including \( \exp + \sqrt{\cdot} \) and \( \exp + (\cdot)^2 \), indicative of broader modeling requirements. Incorporating these domain-specific horizontal dependencies significantly improves the interpretability and practical relevance of symbolic regression outcomes.

\begin{figure}[H]
\begin{center}
\includegraphics[width=6.1in]{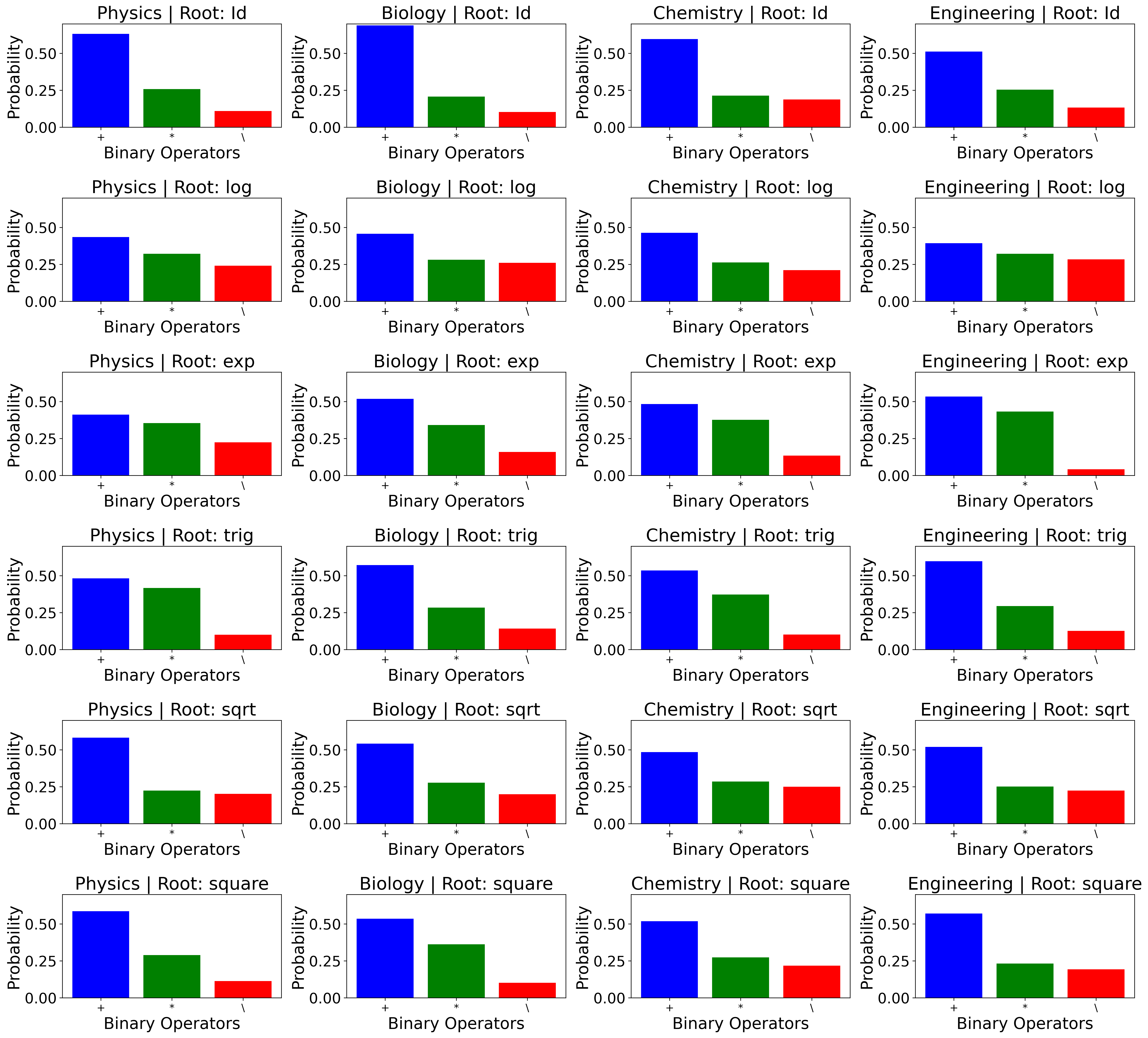}
\end{center}
\caption{Across Physics, Biology, Chemistry, and Engineering, various root unary operators (Id, log, exp, trig, sqrt, square) predominantly connect with addition ($+$) and multiplication ($*$), underscoring their essential roles in aggregating and scaling expressions. However, the specific proportions of these binary operators vary among disciplines, reflecting each field's unique mathematical modeling requirements}
\label{fig: 25}
\end{figure}

Figure~\ref{fig: 25} displays domain-specific distributions of binary operators conditioned on different root node symbols across physics, biology, chemistry, and engineering. The results reveal distinct and consistent operator preferences for each domain. Across all fields, the addition operator ($+$) predominates, reflecting a universal tendency to combine terms directly without transformations. Physics and engineering demonstrate substantial use of multiplication ($\times$), indicative of complex interactions and layered dynamics commonly modeled in these disciplines. Biology and chemistry show relatively balanced usage of addition and multiplication, while division ($\div$) consistently exhibits the lowest frequency, likely due to its numerical instability and less frequent natural occurrence in models across disciplines. These clear patterns underline the domain-specific structures inherent in mathematical expressions, supporting the integration of tailored prior knowledge into the symbolic regression framework to enhance both model interpretability and predictive accuracy.

\section{Methods}
In this section, we present a reinforcement learning-based approach to identify the skeleton of mathematical expressions and subsequently optimize the associated coefficients to finalize the expression learning. Although the training concept is similar to the FEX \cite{liang2022finite} approach, the proposed method is innovative, featuring substantially different elements such as novel tree structures, optimization regularization, and domain-specific prior knowledge. Given a fixed tree structure $\mathcal{T}$ with $n_{\mathcal{T}}$ nodes, FEX aims to identify an expression with finitely many operators to fit given data $\{X, y\}$ by solving $\min_{e, \theta} \mathcal{L}(g(X; \mathcal{T}, e, \theta))$, where $\mathcal{L}$ is a functional quantifying how good an expression $g(X; \mathcal{T}, e, \theta)$ is to fit data, $e$ is the sequence of operators to form $g$, and $\theta = \{\alpha, \beta, \gamma\}$ represents the learnable parameters in $\mathcal{T}$ to form $g$. The expression $g(X; \mathcal{T}, e, \theta)$ is formed by the chosen operators and parameters within the tree structure. This problem is addressed by alternating between optimizing $e$ using reinforcement learning (e.g., policy gradients) and optimizing $\theta$ using gradient-based methods (e.g., Adam, BFGS).

\subsection{Agent}
In this section, we introduce a novel tree-structured recurrent neural network (RNN) designed to function as our agent. As illustrated in Figure 4, this structure enables efficient exploration and representation of complex expressions by capturing hierarchical relationships within the expression tree.
\begin{figure}[ht]
\begin{center}
\includegraphics[width=4.7in]{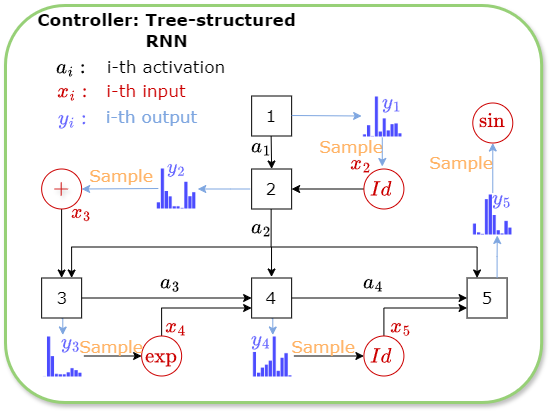}
\end{center}
\caption{Tree-structured RNNs for Symbolic regression}
\label{fig: 7}
\end{figure} 
In this tree-structured RNN, each output $y^i$ represents a categorical probability distribution, indicating the likelihood of selecting various operators for the $i-$th node. The operators $x^i$ are then sampled based on the probabilities provided by $y^{i-1}$, the output from the preceding node. The activations $a^{i}$ propagate through the structure, passing from parent nodes to all child nodes, or horizontally between sibling nodes. This setup allows the model to capture and learn hierarchical dependencies among nodes, reflecting the structured relationships inherent in mathematical expressions.

The key advantage of this structure:
\begin{itemize}
    \item [$\square$]\textbf{Preservation of Structured Information: } This tree-structured RNN is designed to maintain the hierarchical relationships inherent in mathematical expressions. By allowing activations to flow from parent nodes to child nodes and horizontally between sibling nodes, the model preserves the natural structure of expressions. Each node not only receives information from its parent but also shares information with its siblings, enabling the RNN to capture dependencies at multiple levels. This structure aligns closely with the nested and layered nature of mathematical expressions, ensuring that important contextual relationships are retained throughout the network.

    \item[$\square$] \textbf{Efficient Information Flow: } The hidden layer output of a parent node is propagated to all its child nodes, reducing the number of RNN blocks required compared to traditional binary tree methods.
\end{itemize}

\textbf{Remark.}  
In the tree-structured RNN architecture, certain nodes (blocks) may connect simultaneously to multiple subsequent nodes, each corresponding to sibling nodes in the expression tree. Initially, the distribution emitted by an RNN node, denoted as \(\text{Pr}(\text{child} \mid \text{parent})\), is assumed uniform across all connected child nodes. However, due to sequential sampling of sibling symbols, the selection of an earlier sibling influences the conditional probability distribution of the symbols for subsequent siblings. Formally, this sequential sampling induces dependencies of the form:
\[
\text{Pr}(\text{i-th child} \mid \text{(i-1)-th child},\, \text{parent}),
\]
which allows the tree-structured RNN to explicitly model structured sibling dependencies and thus yields a more accurate and coherent representation of expression trees.


\subsection{KL-divergence: Soft constraint}
In our tree-structured recurrent neural network (RNN) architecture, each node $s_i$ outputs a categorical distribution $y_i$ over the set of possible symbols $\mathcal{S}$, which includes unary operators, binary operators, and variables. To ensure that the learned distributions $y_i$ align with our predefined priors $P^*(s_i \mid S_i, s_p, h)$, we compute the Kullback-Leibler (KL) divergence between the RNN-generated distribution $y_i$ and the prior distribution $P^*(s_i \mid S_i, s_p, h)$ for each node $s_i$.

The KL divergence for node \( s_i \) is defined as:
\[
\text{KL}\left(P^*(s_i \mid S_i, s_p, h) \parallel y_i\right) = \sum_{s \in \mathcal{S}} P^*(s \mid S_i, s_p, h) \log\left(\frac{P^*(s \mid S_i, s_p, h)}{y_i(s)}\right)
\]

To aggregate the KL divergences computed for each node within the expression tree, we calculate the average KL divergence over all nodes:

\[
\text{KL}_{\text{avg}} = \frac{1}{N} \sum_{i=0}^{N-1} \text{KL}\left(P^*(s_i \mid S_i, s_p, h) \parallel y_i\right)
\]
Where $N$ is the total number of nodes in the expression tree.

\subsection{Formula Rule: Hard constraint}
We define a set of operator combinations that are prohibited from appearing along the same path within an expression tree. As discussed in Section 2, we observe that many operator combinations are absent from the collected subsequences. This absence may result from various factors: these combinations might violate established symbol rules \cite{petersen2019deep}, lead to numerical instability, or simply be uncommon in the specific domain or due to insufficient data.

We formalize this set as $\text{HConstraint} = \{HC_{1}, HC_{2}\}$:
\begin{itemize}
    \item [$\square$] $HC_{1}$: Represents combinations that violate symbolic rules or result in numerical instability, as identified in prior research. These combinations are strictly prohibited and are excluded from the sampling process.

    \item [$\square$] $HC_{2}$: Represents combinations that rarely occur. Although they are not commonly observed, we assign them a very small probability, $\epsilon$, and include them in the set of soft constraints. This design promotes model exploration, enabling the potential discovery of novel physical laws.
\end{itemize}
For the operators combinations in hard constraint, we simply use the method in \cite{petersen2019deep}, zero-out the probability during sampling. 

By categorizing constraints in this way, we ensure that our model adheres to known rules while still allowing flexibility for exploration. This approach balances enforcing known constraints with maintaining a level of uncertainty, enabling the model to explore new combinations that might reveal novel insights.

\subsection{Reward}

The reward for an operator sequence $e=\{s_{0}, s_{1},...,s_{N-1}\}$, denoted as $R(e)$, is defined as:

\[
R(e) := \frac{1}{1 + \mathcal{L}(e)},
\]
$\mathcal{L}(e) = \min_{\theta} \text{NRMSE}$. This reward $R(e)$ ranges between 0 and 1, where lower values of $R(e)$ result in rewards closer to 1, indicating a better fit to the target equation. Conversely, higher $\mathcal{L}(e)$ values lead to lower reward. 

\subsection{Agent update}
The agent is updated using a basic policy gradient method with a KL-divergence regularization term to regulate the exploration. This regularization controls the distance between the learned policy and the domain-specific prior distribution. Detailed implementation procedures including algorithmic steps and optimization strategies, are provided below:

The agent $\mathcal{A}_{\Psi}$ is implemented as a recurrent neural network (RNN) with parameters $\Psi$. The KL-Regularized training objective of the agent trades off maximizing returns with staying close to the sequences associated with our symbolic prior. This objective is formulated as:

\[
\mathcal{J}(\Psi) = \mathbb{E}_{e \sim \mathcal{A}_{\Psi}}\Big[R(e)-\ell\frac{1}{N} \sum_{i=0}^{N-1} \text{KL}\left(P^*(s_i \mid S_i, s_p, h) \parallel y_i\right)\Big],
\]
where $y_i$ is the $i$-th output of $\mathcal{A}_{\Psi}$, $\ell$ is the hyperparameter.


To optimize the controller $\mathcal{A}_{\Psi}$, we employ a policy gradient-based updating method in reinforcement learning (RL). In practice, we compute an approximation of this gradient using a batch of $k$ sampled operator sequences $e^{(1)}, e^{(2)}, \dots, e^{(k)}$ as follows:

\[
\nabla_{\Psi} \mathcal{J}(\Psi) \approx \frac{1}{k} \sum_{n=1}^{k} R(e^{(n)}) \sum_{i=0}^{N-1} \Big[\nabla_{\Psi} \log (y_i^{(n)}) - \dfrac{\ell}{N}\nabla_{\Psi}\text{KL}\left(P^*(s_i \mid S_i, s_p, h) \parallel y_i\right)\Big].
\]

To update the parameters $\Psi$ of the agent, we use the gradient ascent method with a learning rate $\eta$:

\[
\Psi \leftarrow \Psi + \eta (\nabla_{\Psi} \mathcal{J}(\Psi)).
\]

The goal of the objective function $\mathcal{J}(\Psi)$ is to improve the average reward of the sampled operator sequences. To enhance the probability of obtaining the best equation expression, we modify the objective function using the risk-seeking policy gradient approach:

\[
\mathcal{J}(\Psi) = \mathbb{E}_{e \sim \mathcal{A}_{\Psi}}[R(e) \cdot \mathbb{I}(R(e) \geq R_{\Psi})],
\]

where \(R_{\Psi}\) represents the \((1 - \alpha)\)-quantile of the reward distribution generated by $\mathcal{A}_{\Psi}$, and $\alpha \in [0, 1]$. The gradient computation is updated as:

\[
\nabla_{\Psi} \mathcal{J}(\Psi) \approx \frac{1}{N} \sum_{n=1}^{N} \left( R(e^{(n)}) - \hat{R}_{\alpha} \right) \mathbb{I}(R(e^{(n)}) \geq \hat{R}_{\alpha}) \sum_{i=1}^{k} \nabla_{\Psi} \log (y_i^{(n)}),
\]

where $\hat{R}_{\alpha}$ is an estimate of $R_{\alpha}$ based on the sampled operator sequences. This adjustment improves the convergence of the controller $\mathcal{A}_{\Psi}$ by focusing on higher-reward sequences. To obtain the final symbolic expression generated by our tree-structure RNN, we employ a FEX-based algorithm. 
\begin{algorithm}[H]
\caption{Regularized FEX with tree structure RNNs}\label{alg:cap}
\textbf{Input: } Data $X$, a tree structre $\mathcal{T}$, search loop iteration $T$, coarse-tune iterations $T_{1}$(using Adam) and $T_{2}$(using BFGS), fine-tune iteration $T_{3}$, pool size $K$ and batch size $N$.\\
\textbf{Output: The expression $(\mathcal{T}^*, \theta^*)$}
\begin{algorithmic}[1]
\State Initialize an agent $\mathcal{A}_{\Psi}$ for the tree $\mathcal{T}$ and an empty $\mathcal{P}$
\For{$t$ from $1$ to $T$}
    \State Sample $N$ sequences $\{e^{(1)},...,e^{(N)}\}$ from the agent.
    \For{$n$ from 1 to $N$}
        \State Optimize the NRMSE using both coarse-tune iterations $T_{1}+T_{2}$
        \State Compute the reward for each sequence.
        \State Compute KL divergence
        \If{$e^{n}$ belongs to the top-$K$}
            \State $\mathcal{P}$$.$append($e^{n}$)
        \EndIf
    \EndFor
    \State $g \leftarrow\frac{1}{N} \sum_{n=1}^{N} \left( R(e^{(n)}) - \hat{R}_{\alpha} \right) \mathbb{I}(R(e^{(n)}) \geq \hat{R}_{\alpha}) \sum_{i=1}^{k} \nabla_{\Psi} \log (y_i^{(n)})$
    \State $g_{KL} \leftarrow  -\frac{\ell}{N} \sum_{n=1}^{N}\sum_{i=0}^{|\mathcal{T}|-1} \dfrac{1}{|\mathcal{T}|}\nabla_{\Psi}\text{KL}\left(P^*(s_i \mid S_i, s_p, h) \parallel y_i\right)\Big]$
    \State $\Psi \leftarrow \eta (g+g_{KL})$
\EndFor
\For{$e$ in $\mathcal{P}$}
    \State Fine-tune NRMSE using $T_{3}$ iterations
\EndFor\\
\Return{the expression with smallest fine-tune error}
\end{algorithmic}
\end{algorithm}

\subsection{Dynamic Scheduling of KL Divergence Regularization to Balance Prior Influence}

A potential concern with the inclusion of priors is that they may introduce bias, forcing the model to remain overly close to the prior categorical distribution throughout training. To address this, we employ a dynamic scheduling strategy for the KL divergence regularization term. Specifically, we adapt the hyperparameter $\ell$ to gradually decay as training progresses. 

In the early stages of training, a larger $\ell$ ensures that the model leverages prior knowledge to accelerate convergence and stabilize the search process. As training continues, $\ell$ is gradually reduced, allowing the model to rely more on the observed data and explore solutions beyond the initial priors. This balance mitigates the risk of excessive prior influence while preserving the benefits of guided exploration.

The decay of $\ell$ can be implemented using an exponential schedule:
\[
\beta(t) = \ell_0 \cdot \exp(-\lambda_{d} t),
\]
where $\ell_0$ is the initial regularization weight, $\lambda_{d}$ is the decay rate, and $t$ represents the training epoch. This dynamic adjustment ensures that the model transitions smoothly from a prior-dominated phase to a data-driven optimization phase, ultimately improving flexibility and generalization. The initial value of the KL divergence regularization term \(\ell_0\) and the decay rate \(\lambda\) for the dynamic scheduling were determined using grid search over a predefined range, ensuring optimal trade-offs between prior alignment and model flexibility.

\section{Experiments}
In this section, we conduct a series of experiments to evaluate the effectiveness, performance, and generalization capabilities of our proposed method. The experiments are designed to comprehensively assess the method across multiple domains, compare it with existing approaches, and investigate the impact of incorporating domain-aware symbolic priors. We begin with a Benchmark Test to validate our method on standardized symbolic expressions, followed by a Comparative Analysis Across Domains to demonstrate its robustness and versatility. 

\subsection{Benchmark Test}
In this section, we evaluate the performance and generalization capabilities of our proposed method on a standardized set of benchmark expressions. We construct the benchmark test to include symbolic expressions across various domains and complexities. This ensures a comprehensive comparison between our method and existing approaches. we employ two well-established benchmarks to evaluate the performance of our proposed method across distinct domains:

\begin{itemize}
    \item \textbf{Feynman Benchmark for Physics}: A collection of symbolic expressions derived from the \textit{Feynman Lectures on Physics}, covering fundamental laws across mechanics, electromagnetism, thermodynamics, and quantum mechanics. These expressions are known for their dimensional consistency and real-world applicability, making them a rigorous test for symbolic regression methods in the physics domain\cite{la2021contemporary}.
    
    \item \textbf{ODEbase for Biology}: The \textit{ODEbase} database serves as a benchmark for biological expressions. It consists of mathematical models, primarily in the form of ordinary differential equations (ODEs), that are widely used to describe biological processes, such as population dynamics, biochemical reactions, and cellular behaviors. From ODEbase, we extract representative expressions to evaluate the ability of our method to build accurate models for biological systems.\cite{LuedersSturmRadulescu:22}
\end{itemize}

We meticulously adhered to the established protocol delineated in SRBench by La Cava et al. (2021)\cite{la2021contemporary} for evaluating the performance of our method across the two selected benchmarks: Feynman Benchmark for Physics and ODEbase for Biology.

\begin{itemize}
    \item For the Feynman Benchmark, our algorithm was tasked with identifying symbolic expressions that fit $10,000$ data points corresponding to each Feynman benchmark equation. Exact symbolic recovery was determined by verifying that the difference between the generated expression and the target expression simplified to a constant.
    
    \item For the ODEbase Benchmark, we extracted \textbf{200 representative expressions} describing biological systems modeled by ordinary differential equations (ODEs). The recovery criteria and symbolic simplification checks were identical to those applied in the Feynman Benchmark.
\end{itemize}
To ensure robustness, each experiment was repeated \textbf{10 times} with different random seeds, and recovery rates were averaged across these trials.

\noindent \textbf{Noise Levels}:  
In alignment with benchmark practices, experiments were conducted across four noise levels: $0$, $0.01$, $0.05$, $0.07$ and $0.1$.

\noindent \textbf{Data Usage}:  
Each benchmark problem was defined by:
\begin{itemize}
    \item A \textbf{ground truth expression} for symbolic recovery validation.
    \item A \textbf{training dataset} used to compute the reward for candidate expressions during optimization.
    \item A \textbf{test dataset} used to evaluate the final candidate expression at the end of training.
\end{itemize}

To ensure the robustness and reliability of the results, each experiment was repeated using 100 different random seeds for every benchmark expression. The recovery rate was determined as the proportion of runs in which the algorithm successfully identified the target expression, following the exact symbolic recovery criteria outlined earlier. By applying this rigorous evaluation protocol, we ensured a fair and comprehensive assessment of our method's performance across the physics and biology benchmarks.
\begin{figure}[h]
\begin{center}
\includegraphics[width=6.5in]{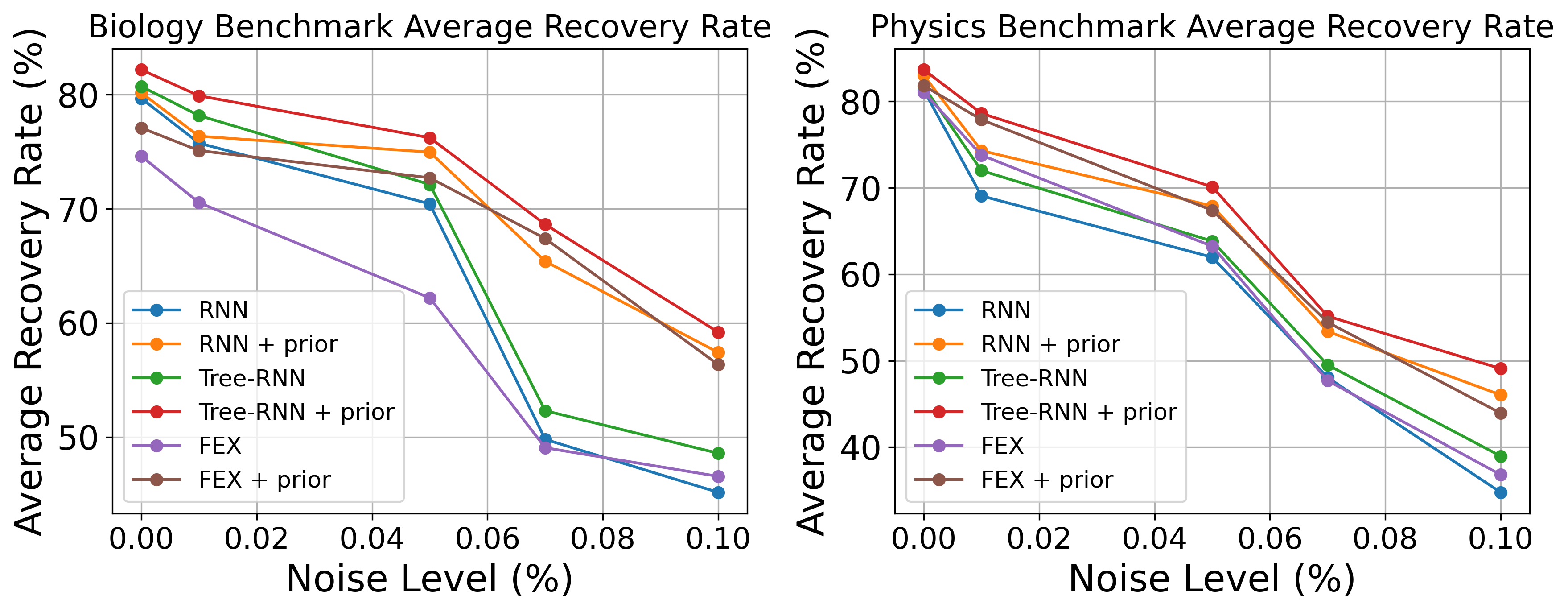}
\end{center}
\caption{Comparison of recovery rates across varying noise levels for Biology and Physics benchmarks. Results are shown for six methods: RNN, RNN + prior, Tree-RNN, Tree-RNN + prior, FEX, and FEX + prior.}
\label{fig:recovery_rate}
\end{figure}

The results presented in Figure~\ref{fig:recovery_rate} highlight the significant impact of incorporating prior knowledge and utilizing hierarchical structures in symbolic regression tasks under varying noise conditions.

First, methods that integrate prior knowledge (\textit{RNN + prior}, \textit{Tree-RNN + prior}, and \textit{FEX + prior}) demonstrate consistently higher recovery rates across all noise levels compared to their counterparts without priors. This trend is particularly evident in both benchmarks, where the inclusion of priors effectively mitigates the performance degradation caused by increasing noise. Prior knowledge serves as a regularization mechanism, constraining the solution space and enabling the models to generalize more robustly in noisy environments.

Second, among the tested methods, \textit{Tree-RNN} and \textit{Tree-RNN + prior} exhibit superior performance, particularly at moderate and high noise levels. The hierarchical representation in Tree-RNN allows for efficient modeling of complex expressions, reducing search depth and improving recovery accuracy. Notably, the \textit{Tree-RNN + prior} method achieves the highest recovery rates across all conditions, underscoring the complementary benefits of combining structural efficiency with domain-specific priors.

In contrast, traditional methods such as \textit{RNN} and \textit{FEX} show a sharp decline in performance as noise levels increase, indicating their sensitivity to noisy data. The results confirm that the integration of priors and hierarchical tree structures not only enhances recovery robustness but also improves the scalability of symbolic regression methods to real-world noisy datasets.

\subsection{Comparative Analysis Across Domains}
In this section, we choose four expressions from four distinct domains to conduct a comparative analysis of the following methods: FEX, FEX with priors, RL + RNN, RL + RNN with priors, RL + tree-structured RNN, and RL + tree-structured RNN with priors. The detailed descriptions of the six expressions utilized in this experiment are provided in ~\ref{app:D}.

Learning parameters for the first two problems are: learning rate $0.003$, batch size $1000$, risk factor is $0.05$, KL divergence parameter is $0.5$. For the other two: learning rate $0.001$, batch size $1000$, risk factor is $0.05$, KL divergence parameter is $0.35$.
\begin{figure}[h]
\begin{center}
\includegraphics[width=6.8in]{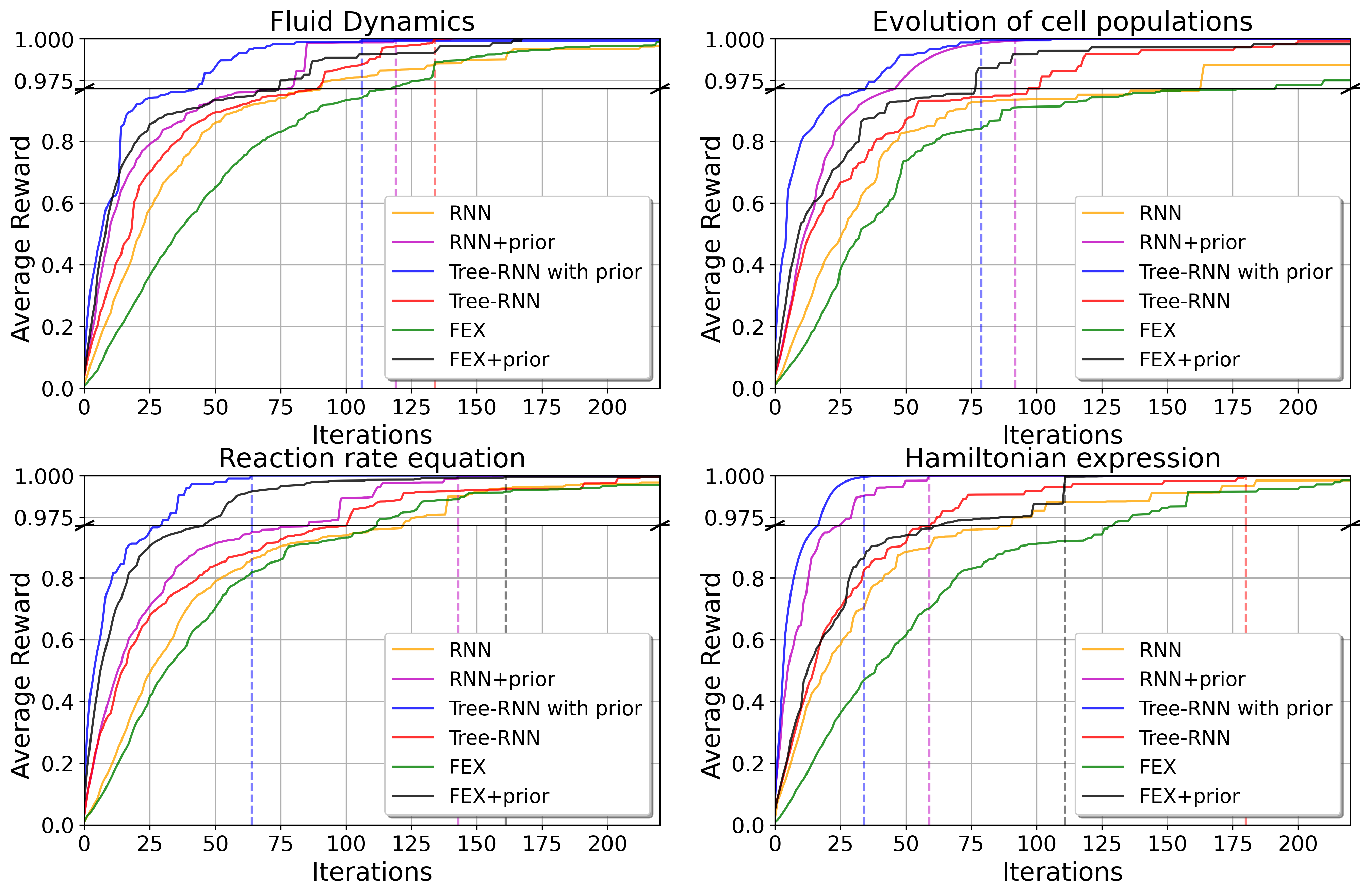}
\end{center}
\caption{
Results across four different scenarios:
(\textbf{Top-left})~Fluid Dynamics,
(\textbf{Top-right})~Evolution of cell populations,
(\textbf{Bottom-left})~Reaction rate equation, and
(\textbf{Bottom-right})~Hamiltonian expression.
Each subfigure shows results for six methods:
\emph{RNN}, \emph{RNN+prior}, \emph{Tree-RNN}, \emph{Tree-RNN+prior}, \emph{FEX}, and \emph{FEX+prior}.
}
\label{fig:recovery_rate}
\end{figure}

Based on experiments, we can draw several important conclusions about the effectiveness of using prior knowledge and tree-structured RNNs:
\begin{itemize}
    \item [$\bullet$] \textbf{Effectiveness of Priors and Tree-Structured RNNs: } The incorporation of domain-specific priors and tree-structured RNNs significantly enhances learning efficiency. Both ``Tree-RNN with prior" and ``RNN-prior" converge more quickly to optimal policies compared to other methods, demonstrating the advantage of leveraging prior knowledge and hierarchical architectures. For instance, the Hamiltonian expression, characterized by numerous additive terms, poses challenges for traditional RNNs. The tree-structured RNN reduces the required network depth for such additive structures, while the prior categorical distribution equips the model with domain-specific insights. This combination accelerates convergence and enables efficient identification of high-quality solutions.
    \item [$\bullet$] \textbf{Variability in Prior Impact: } The fourth figure (Fluid Dynamics) reveals a potential drawback of using priors. Here, the methods incorporating prior knowledge (``Tree-RNN with prior" and ``RNN-prior") do not outperform the other methods by a large margin. This suggests that priors can introduce biases that may not always align well with certain complex expressions, thereby limiting their effectiveness.
\end{itemize}

In general, our results demonstrate that combining domain-specific priors with a tree-structured RNN agent can significantly enhance the learning of complex functions. However, as illustrated in the fourth figure, the incorporation of priors may sometimes introduce biases, leading to suboptimal performance in certain cases. This variability in effectiveness highlights the need for careful consideration and selection of priors to match the characteristics of the problem domain.

\section{Conclusion}
We find that combining domain-specific priors with our tree-structured RNN agent quickly results in an effective policy. Learning from expressions across various fields has provided valuable insights for future research. However, our approach is sensitive to the prior categorical distribution, making bias a challenge despite careful data collection.

The prior for each domain consists of two parts: a ``behavior prior" shared across all fields, and a domain-specific component. This is similar to the multitask problem in reinforcement learning. In future work, we plan to optimize both the domain-specific and ``behavior" priors during training, aiming to uncover intriguing and interesting results.

\section*{Acknowledgement}
Haizhao Yang were partially supported by the US National Science Foundation under awards DMS-2244988, DMS-2206333, the Office of Naval Research Award N00014-23-1-2007, and the DARPA D24AP00325.

\newpage
 \bibliographystyle{elsarticle-num} 
 \bibliography{cas-refs}





\newpage
\appendix

\section{Descriptions of Expressions}
\label{app:D}
In physics, we compare different SR method to recover Hamiltonian expression. The Hamiltonian $H$ for a nuclear system with a simplified model involving three momentum variables $p_1, p_2, p_3$ is given by:

\[
H = \frac{\hat{A} - 1}{\hat{A}} \sum_{i=1}^{3} \frac{p_i^2}{2m_N} - \frac{1}{m_N \hat{A}} \sum_{i < j}^{3} p_i \cdot p_j + \sum_{i < j}^{3} V_{ij}.
\]

Where $m_N$ (Nucleon Mass) represents the average mass of a nucleon (either a proton or a neutron) in the nuclear system. It is used in kinetic energy calculations. The average nucleon mass simplifies computations, as the system contains multiple nucleons. $\hat{A}$(Particle-Number Operator) is an operator representing the total number of nucleons (particles) in the system. In the given context, $\hat{A}$ can be treated as the scalar number of nucleons, often denoted by $A$. The operator form is used in many-body physics to handle systems with varying particle numbers.
$p_i$(Momentum) represents the momentum of the $i$-th nucleon. In this simplified model, only three momentum variables ($p_1, p_2, p_3$) are considered. $V_{ij}$ (Two-Body Potential) represents the interaction energy between nucleons $i$ and $j$ . This term accounts for forces between pairs of nucleons and can take various forms depending on the nature of the interaction. We use a simplified form, such as \( V_{ij} = \frac{g}{r_{ij}} \), where \( g \) is a constant.
Given these variables and terms, the simplified Hamiltonian expression for the system involving three momentum variables (\( p_1, p_2, p_3 \)) is:

\[
H = \frac{\hat{A} - 1}{\hat{A}} \sum_{i=1}^{3} \frac{p_i^2}{2m_N} - \frac{1}{m_N \hat{A}} \sum_{i < j}^{3} p_i \cdot p_j + \sum_{i < j}^{3} \frac{g}{r_{ij}},
\]
We set $A^{hat} = 2.0$
$m_N = 1.5$, $g = 0.8$.

\textbf{In biology}, we always describe the evolution of four distinct cell populations within a tumor microenvironment during the course of treatment. These populations include two sub-populations of tumor cells and two types of interacting cells (CAR T-cells and bystander cells). The model uses a system of differential equations to capture the dynamics of these populations.
   The simplified form of Equation (4) now looks like:
   \[
   \frac{dB}{dt} = b - \gamma_B B - \mu_B \log \left( \frac{B + C}{K_2} \right) + \frac{\left( d_B + s \left( \frac{B}{T_s} \right)^2 \right)^2}{k + \left( d_B + s \left( \frac{B}{T_s} \right)^2 \right)^2} B - \omega_B B (T_s + T_r).
   \]

Where \(T_s\) and \(T_r\) are variables representing the tumor sub-populations. $B$ is the bystander cell population. $C$ is the CAR T-cell population. We set $ b=0.5, gamma_B=0.1, mu_B=0.3, K2=1.0, d_B=0.05, s=2.0, k=0.8, omega_B=0.2$.

\textbf{In chemistry:} 
Reaction Rate Equation for $n = 3$:

Given three substrates (\( S_1, S_2, S_3 \)) and an inhibitor \( I \), the equation can be written as:

\[
v = \frac{V_{\max} \cdot [S_1] \cdot [S_2] \cdot [S_3]}{\left(K_m + [S_1] + [S_2] + [S_3]\right) \left(1 + \frac{[I]}{K_i}\right)}
\]

We keep \( V_{\max} = 1.0 \), \( K_m = 0.5 \), and \( K_i = 0.3 \). You can modify these parameters as needed.
\begin{itemize}
    \item [$\square$] Random Concentrations: We generate random concentrations for three substrates (\( S_1, S_2, S_3 \)) and one inhibitor (\( I \)) within specified ranges.
    \item [$\square$] Reaction Rate Calculation: The reaction rate is computed using the updated equation that involves three substrates.
\end{itemize}

In Engineering, A deep function in the context of engineering can be a composition of multiple nested unary and binary operators, often found in fields like control systems, fluid dynamics, signal processing, or structural engineering. The more nested or ``deep" the operations, the more challenging it becomes for symbolic regression to approximate. 

Here's an example of a complicated ``deep" function inspired by fluid dynamics and turbulence modeling. This function includes multiple layers of unary operations such as logarithms, trigonometric functions, and nested square roots:

\[
f(x) = \log\left(\alpha \sqrt{x} + \beta \sin(\gamma x + \delta) \right) + \frac{\epsilon}{\cos\left(\eta \sqrt{x} + \theta \log(x) \right) + \zeta \exp\left(-\lambda x^2\right)}.
\]
\(\alpha, \beta, \gamma, \delta, \epsilon, \eta, \theta, \zeta, \lambda\) are coefficients that control the function's shape and behavior. We set coefficients $\alpha=1.2, \beta=0.8, \gamma=2.0, \delta=0.5,
    \epsilon=0.1, \eta=1.5, \theta=0.3, \zeta=0.05, \lambda_=0.01$.

\end{document}